\pdfoutput=1

\documentclass[11pt]{article}

\usepackage[]{acl}

\usepackage{times}
\usepackage{latexsym}
\usepackage{multicol}
\usepackage{multirow}
\usepackage{graphicx}
\usepackage[T1]{fontenc}

\usepackage{algorithm}
\usepackage{algorithmic}
\usepackage{makecell}
\usepackage{pgfplots}

\usepackage[utf8]{inputenc}

\usepackage{microtype}

\usepackage{inconsolata}

\title{GADFA: Generator-Assisted Decision-Focused Approach for Opinion Expressing Timing Identification}

\author{Chung-Chi Chen,\textsuperscript{1} Hiroya Takamura,\textsuperscript{1} Ichiro Kobayashi,\textsuperscript{2}  Yusuke Miyao,\textsuperscript{3} Hsin-Hsi Chen\textsuperscript{4}
\\
 \textsuperscript{1} Artificial Intelligence Research Center, AIST, Japan \\
\textsuperscript{2} Ochanomizu University, Japan \\
 \textsuperscript{3} University of Tokyo, Japan \\
 \textsuperscript{4} Department of Computer Science and Information Engineering, \\ 
   National Taiwan University, Taiwan \\
   \texttt{c.c.chen@acm.org, takamura.hiroya@aist.go.jp,}\\ \texttt{koba@is.ocha.ac.jp, yusuke@is.s.u-tokyo.ac.jp, hhchen@ntu.edu.tw}\\
}

\begin{document}
\maketitle
\begin{abstract}
The advancement of text generation models has granted us the capability to produce coherent and convincing text on demand. Yet, in real-life circumstances, individuals do not continuously generate text or voice their opinions. For instance, consumers pen product reviews after weighing the merits and demerits of a product, and professional analysts issue reports following significant news releases. In essence, opinion expression is typically prompted by particular reasons or signals. Despite long-standing developments in opinion mining, the appropriate timing for expressing an opinion remains largely unexplored. To address this deficit, our study introduces an innovative task - the identification of news-triggered opinion expressing timing. We ground this task in the actions of professional stock analysts and develop a novel dataset for investigation. Our approach is decision-focused, leveraging text generation models to steer the classification model, thus enhancing overall performance. Our experimental findings demonstrate that the text generated by our model contributes fresh insights from various angles, effectively aiding in identifying the optimal timing for opinion expression.
\end{abstract}

\section{Introduction}

Opinion mining has been a popular topic for a long time \cite{liu2012sentiment}. We are now able to perform well in sentiment analysis \cite{wu-shi-2022-adversarial}, aspect-based sentiment analysis \cite{ling-etal-2022-vision,chen-etal-2022-discrete}, opinion helpfulness estimation \cite{diaz2018modeling}, and several opinion understanding tasks \cite{shi-etal-2022-effective}. With the development of large pre-trained language models, we have also made significant improvements in synthesizing opinions and arguments \cite{wachsmuth-etal-2018-argumentation,hua-etal-2019-argument-generation,schiller-etal-2021-aspect}. Although there are many discussions on how opinions are generated and what should be generated, few studies pay attention to when we should generate opinions. Since models can generate text based on given input at any time, timing becomes an important issue. For example, there are many news articles every day, but professional analysts do not write comments on every article and only release reports a few times per year. Following this line of thought, this paper proposes a novel task: identifying the timing of expressing opinions. The aim of this task is to learn when is the proper timing to express opinions.

Professionals' opinions, attitudes, and behaviors are important for the masses' decision-making and the future of the field, market, country, and so on. For example, the opinions of the professionals in the Centers for Disease Control and Prevention (CDC) will influence people's attitudes toward COVID-19 in the past three years, and politicians' attitudes will influence the country-level and even global-level political situations. In the financial market, the opinions and behaviors of the professionals are shown to be related to different market properties \cite{hirst1995investor,niehaus2010impact,kothari2016analysts,kim2022firm}. Given the importance of professionals' opinions, attitudes, and behaviors, this paper aims to forecast professional stock analysts' behaviors based on the latest publicly-available information, i.e., news.

When the latest news was published, the first decision that analysts will make is whether they should write a report to update their view and explain it to their customers. That is, if the news does not matter to the company's operation or stock price, analysts will not release reports to share their opinions. Because investors will react to the released reports \cite{hirst1995investor,niehaus2010impact}, we argue that this is an essential task in analysts' behavior modeling, and it is also an ideal data for exploring the proposed timing identification task. To the best of our knowledge, this paper is the first work to explore the news-triggered opinion expressing timing identification task.

Training classification models with all news as input is an intuitive way to address the proposed task. However, given the length limitation of models' input, generating news summarization before training a classifier is the other rational direction for news-triggered classification tasks. Different from the above directions, this paper attempts to answer the question of whether we can improve the performances by adding an opinion generator in the loop. The rationale of our design for this approach is that we want to let models mimic the decision process of professionals. Instead of directly deciding whether to write a report after reading the news, professionals will first form their opinions toward the given news, and then further decide whether to share this opinion with their customers. Therefore, we propose a Generator-Assisted Decision-Focused Approach (GADFA), which first trains a generator to generate analysis based on each news, and further uses the generated opinions and the news as models' input. The important difference between the generated opinions and summarizations is that opinions contain subjective information which is not included in the original articles. In contrast, summarization is just to rewrite and shorten the given news.

In sum, this paper makes the following contributions:
(1) We propose a novel task: news-triggered opinion expressing timing identification.
(2) We design a decision-focused approach for enhancing the performances.
(3) We provide in-depth discussions on the influence of using different generators and using cross-generators in the proposed approach, and also analyze the generated text from several aspects.

\section{Related Work}
As it is relatively easy to collect textual data, such as tweets or news articles, and align them with market prices, numerous datasets have been developed for predicting market information, including price movement prediction~\cite{xu-cohen-2018-stock,li2020modeling} and volatility forecasting~\cite{qin-yang-2019-say,li2020maec}.
In our belief, short-term price movement follows the random walk hypothesis~\cite{fama1995random}, and several asset pricing models have adopted this concept to model asset price movement, such as the Black–Scholes model~\cite{black1973pricing} for option pricing. 
Building upon this notion, we argue that learning to make professional decisions is a more tangible direction. 
Thus, this paper aligns news articles with professionals' behavior, specifically whether professional analysts will release reports after the news is published. 
The proposed task has several downstream applications.
For example, choosing the timing to share opinions is an important task when constructing an AI analyst because we expect the AI analyst to only share important information instead of generating numerous unnecessary explanations.  
In other words, rather than sharing generated text continuously, the AI analyst also needs to select the timing of expressing opinions. 

Analysts' behavior has been a long-standing topic in financial literature. 
Some studies focus on analyzing reports and market reactions.
For instance, \citeauthor{devos2015stock} (\citeyear{devos2015stock}) discuss the market response to changes in analysts' views, indicating that analysts' view changes are informative for investment, particularly for stocks with less transparency. 
\citeauthor{hsieh2016analyst} (\citeyear{hsieh2016analyst}) examine the readability of reports and its impact on stock returns, finding that readability plays a significant role in eliciting positive reactions from the market.  
Others attempt cross-document analysis. 
\citeauthor{conrad2006analyst} (\citeyear{conrad2006analyst}) explore the relationship between analyst recommendations and major news. 
\citeauthor{keith-stent-2019-modeling} (\citeyear{keith-stent-2019-modeling}) model changes in analysts' views based on pragmatic and semantic features of earnings calls. 
To summarize our survey, there are existing discussions on analysts' view changes, but few studies have focused on the timing of opinion expression. 
Furthermore, there is currently no publicly-available dataset for investigating the expressing timing of analysts' opinions. 
The dataset proposed in this paper is the first of its kind released for such a task.

\section{Dataset}
\subsection{Task Design}
While numerous new events occur daily in the financial market, originating from various information sources, most studies assume that news articles contain the most up-to-date information and also summarize information from other documents such as financial reports or company meetings. Additionally, news articles may report on popular discussion threads from social media platforms. Therefore, we have chosen news articles as the primary source of information for our models. 

The proposed task of identifying the timing of opinion expression triggered by news is defined as follows. Given the news related to the target stock from day $t-T$ to $t$, our objective is to predict whether at least one professional analyst will release an analysis report on day $t+1$. In this paper, we conduct experiments for two different values of $T$, namely $T = 0$ and $T = 5$, to discuss the timeliness of the news.

\begin{table}[t]
  \centering
  \small
    \begin{tabular}{l|cc}
          & Release Report & Not Release Report \\
    \hline
    Train & \multicolumn{1}{r}{                2,717 } & \multicolumn{1}{r}{                       2,717 } \\
    Development & \multicolumn{1}{r}{                   322 } & \multicolumn{1}{r}{                          322 } \\
    Test  & \multicolumn{1}{r}{                   325 } & \multicolumn{1}{r}{                          325 } \\
    \hline
    Total & \multicolumn{2}{c}{6,728} \\
    \end{tabular}%
        \caption{Statistics of the dataset.}
  \label{tab:statistics}%
\end{table}%

\begin{table}[t]
  \centering
  \small
    \begin{tabular}{l|rr}
          & \multicolumn{1}{c}{$T = 0$} & \multicolumn{1}{c}{$T = 5$} \\
    \hline
    One News & 3,632 & 3,051 \\
    More Than One News & 3,096 & 3,677 \\
    \end{tabular}%
       \caption{Statistics of the number of news.}
  \label{tab:Statistics of the number of news}%
\end{table}%

\subsection{Dataset Creation Process}
To construct the dataset for the proposed task, we follow a series of steps. Firstly, we download all analysis reports for the Taiwan stock market from Bloomberg Terminal. Additionally, we obtain the Chinese news released by two major financial news vendors, Economic Daily News\footnote{\url{https://money.udn.com/money/index}} and Commercial Times.\footnote{\url{https://ctee.com.tw/}}
Secondly, we align these data based on their respective release times. It is important to note that this dataset covers the period from 2014 to 2020, comprising a total of 401,559 news articles and 40,205 reports.
Thirdly, we filter the instances that have news on day $t$ and at least one report released on day $t+1$. This filtering process results in 3,364 positive instances labeled as "Release Report". 
Instead of randomly selecting negative instances labeled as ``Not Release Report,'' we control the target stock based on the positive instances. This means that the negative instances are selected from the same stock pools as the positive instances. 
The reason for controlling the target stock is that previous studies have shown potential bias towards the target stock in pre-trained models~\cite{chuang-yang-2022-buy} and managers' gender~\cite{sawhney-etal-2021-empirical}. 
Using this approach, we identify negative instances that have news on day $t$ but no report released on day $t+1$.

Table~\ref{tab:statistics} presents the number of instances in the training, development, and test sets. 
Additionally, Table~\ref{tab:Statistics of the number of news} provides statistics on the number of news articles under different settings of $T$ based on the opinion expressing timing detection dataset. 
We observe that when $T$ is set to 0 and 5, 46.16\% and 54.65\% of instances, respectively, have multiple news articles. 
Furthermore, it indicates that 8.64\% ((3,677-3,096)/6,728) of stocks are mentioned on different days during the five-day period.
These results motivate us to explore the timeliness of information and the potential benefits of summarizing multiple news articles in the proposed tasks.

\begin{figure}[t]
  \centering
  \includegraphics[width=8cm]{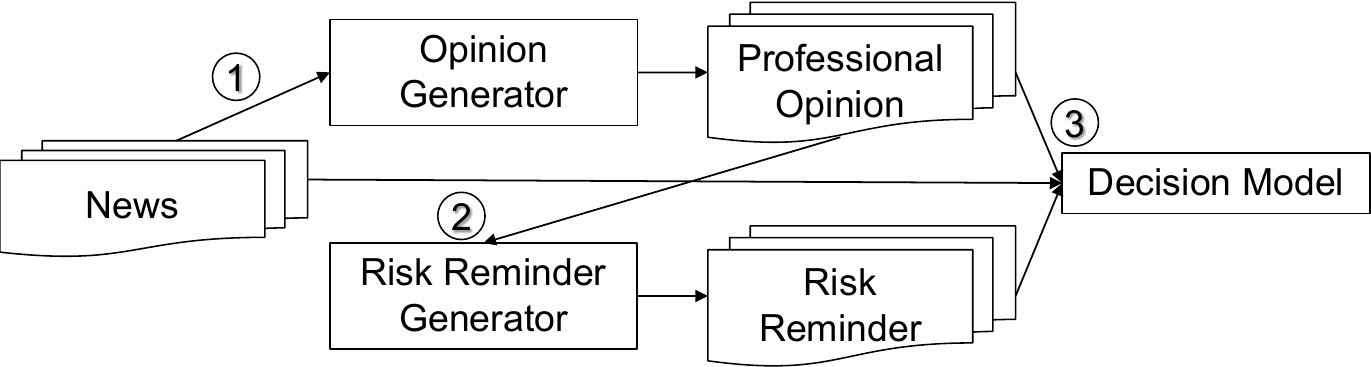}
  \caption{Illustration of the proposed GADFA.}
  \label{fig:Illustration of the proposed GADFA}
\end{figure}

\section{Method}
Figure~\ref{fig:Illustration of the proposed GADFA} illustrates the proposed GADFA. The design of GADFA is inspired by the notion of decision-focused approaches \cite{wilder2019melding,hsu-tan-2021-decision,mandi2022decision}, which involve making several predictions first and then using these predictions as input to an optimization algorithm for generating a decision. We believe that this pipeline approach can also be applied to natural language processing (NLP) applications, as the generated text can contain multiple information pieces from different aspects, such as sentiments and analysis. Considering the behavior model of professionals, we consider generating an opinion on the given news and then making further decisions as an ideal way to perform decision-focused learning for the proposed task.

Therefore, the proposed GADFA consists of three steps. Firstly, the opinion generator generates a professional opinion based on the given news. Table~\ref{tab:example} presents examples of professional opinions, which include subjective views on whether investors should buy or sell the related stocks mentioned in the news. Secondly, the risk reminder generator generates a risk reminder based on the generated opinions. Table~\ref{tab:example} provides examples of risk reminders, which indicate possible risks that may affect the accuracy of the generated opinion. Finally, the decision model combines the given news, generated opinions, and generated risk reminders to determine the appropriate timing for expressing opinions.

As shown in Figure~\ref{fig:Illustration of the proposed GADFA}, the proposed GADFA consists of two generators and one decision model. We first generate a professional opinion based on the given news and then generate a risk reminder based on the generated opinion. The decision model then fuses the news, professional opinion, and risk reminder to make the final decision on whether to release a report or not. For the generators, we select three well-performing pre-trained language models for comparison:

\begin{table}[t]
  \centering
  \resizebox{\columnwidth}{!}{
    \begin{tabular}{lp{15em}}
    \hline
    \multirow{7}[1]{*}{News}  & Key takeaways from Quanta’s 1Q22 analyst call included: 1) 1Q22 GMs dipped to an eight-quarter low due to inferior product mix and inefficient production; 2) weak 2Q22 earnings outlook with both PC demand weakness and supply constraints… \\
    \hline
    \multirow{7}[1]{*}{Professional Opinion} & We believe that QCT could suffer from a potential slowdown in enterprise spending, while China hyperscalers' demand also seems a bit mellowed down in the near term. We forecast 23\%/10\% server revenue growth for Quanta in 2022/23. \\
    \hline
    \multirow{3}[1]{*}{Risk Reminder} & Key downside risks include margin erosion in servers and a potential slowdown of PC demand post Covid-19. \\
    \hline
    \end{tabular}%
    }
        \caption{Examples of the news, professional opinion, and risk reminder.}
  \label{tab:example}%
\end{table}%

\begin{table*}[t]
  \centering
  \small
    \begin{tabular}{l|rrrr|rrrr}
          & \multicolumn{4}{c|}{Opinion Generator} & \multicolumn{4}{c}{Risk Reminder Generator} \\
    \hline
          & \multicolumn{1}{c}{ROUGE-1} & \multicolumn{1}{c}{ROUGE-2} & \multicolumn{1}{c}{ROUGE-L} & \multicolumn{1}{c|}{BERT Score} & \multicolumn{1}{c}{ROUGE-1} & \multicolumn{1}{c}{ROUGE-2} & \multicolumn{1}{c}{ROUGE-L} & \multicolumn{1}{c}{BERT Score} \\
    \hline
    mT5   & 0.1238 & 0.0475 & 0.1195 & 0.6471 & 0.6630 & 0.4878 & 0.6554 & 0.8581 \\
    Pegasus & 0.2098 & 0.0951 & 0.1997 & 0.7034 & \textbf{0.6883} & \textbf{0.5306} & \textbf{0.6801} & \textbf{0.8752} \\
    Mengzi T5 & \textbf{0.2397} & \textbf{0.1150} & \textbf{0.2250} & \textbf{0.7061} & 0.6652 & 0.5130 & 0.6591 & 0.8745 \\
    \end{tabular}%
      \caption{Experimental results of opinion generator and risk reminder generator.}
  \label{tab:Experimental results of generator}%
\end{table*}%

\begin{table*}[t]
  \centering
  \small
 \begin{tabular}{llrr}
    \hline
    \multicolumn{1}{c}{Method} & \multicolumn{1}{c}{Input of Decision Model} & \multicolumn{1}{c}{$T = 0$} & \multicolumn{1}{c}{$T = 5$} \\
    \hline
    CPT   & \multirow{2}[1]{*}{News} & 74.31\% & 70.44\% \\
    \multirow{2}[1]{*}{BERT} &       & 75.64\% & 77.79\% \\
          & News + XL-Sum Abstract & 73.46\% & 77.25\% \\
    \hline
    \multirow{3}[2]{*}{GADFA} & News + mT5 Professional Opinion and Risk Reminder & 77.80\% & 78.28\% \\
          & News + Pegasus Professional Opinion and Risk Reminder & 77.75\% & 78.72\% \\
          & News + Mengzi T5 Professional Opinion and Risk Reminder & \textbf{78.12\%} & \textbf{79.52\%} \\
    \hline
    \end{tabular}%
    \caption{Experimental results of news-triggered opinion expressing timing identification task.}
  \label{tab:Experimental results of timing identification task}%
\end{table*}%

\begin{table}[t]
  \centering
  \small
    \begin{tabular}{lr}
    \multicolumn{1}{c}{Opinion Generator} & \multicolumn{1}{c}{Macro-F1} \\
    \hline
    mT5   & 76.71\% \\
    Pegasus & 78.36\% \\
    Mengzi T5 & 78.34\% \\
    \hline
    mT5 + Pegasus + Mengzi T5 & \textbf{78.54\%} \\
    \end{tabular}%
  \caption{Ablation analysis --- Remove risk reminder.}
  \label{tab:ablation - risk reminder}%
\end{table}%

\begin{itemize}
    \item Multilingual T5 (mT5) \cite{xue-etal-2021-mt5} covers 101 languages and was pre-trained on a Common Crawl-based dataset.
    
    \item Pegasus \cite{zhang2020pegasus}\footnote{The Chinese Pegasus is provided in UER toolkit \cite{zhao2019uer}} is pre-trained with a masked sentence generation task and performs well in summarization tasks.
    
    \item Mengzi T5 \cite{zhang2021mengzi} uses T5 \cite{raffel2020exploring} as a backbone and is pre-trained for Chinese.
\end{itemize}

For the decision model, we use the standard BERT model (\textit{bert-base-chinese}) \cite{devlin-etal-2019-bert} in the proposed approach after comparing it with CPT \cite{shao2021cpt}, which is a tailor-made model for Chinese understanding and generation.

To train the opinion generator, we obtain 2,004 news-opinion pairs from a securities company. Since all opinions are written by professional analysts in the company, we use this data to fine-tune pre-trained language models for generating opinions on the given news. In this experiment, 1,603 (80\%) instances are used for training, and the remaining instances are used for evaluation. To train the risk reminder generator, we extract 1,356 suggestion-reminder pairs from the reports collected from the Bloomberg Terminal. These reports are also written by professional analysts, and the risk reminder is used to indicate possible events that may affect the accuracy of the suggestion. We use this data to fine-tune pre-trained language models for generating risk reminders based on the given professional opinion. We use 80\% of the instances for training and the rest for evaluation.

\section{Experiment}
\subsection{Opinion and Risk Reminder Generators}
To evaluate the results of the generation, we employ ROUGE \cite{lin-2004-rouge} and BERT Score \cite{zhang2019bertscore} as evaluation metrics. Table~\ref{tab:Experimental results of generator} presents the outcomes of different pretrained language models in professional opinion generation and risk reminder generation tasks. In both tasks, the models demonstrate high BERT Scores. Among the three models, Mengzi T5 exhibits the best performance in the professional opinion generation task. We summarize our observations regarding the generated opinions as follows: Firstly, mT5 generates more repeated sentences compared to the other two models, which is the primary reason for its inferior performance. Secondly, Pegasus and Mengzi T5 are capable of producing fluent opinions, albeit with occasional hallucinations such as incorrect numerals. Thirdly, the generated opinions include forward-looking perspectives that are not present in the given news.

In the risk reminder generation task, Pegasus outperforms other models. We observe that both ROUGE and BERT Score yield high scores in the risk reminder generation task. Upon comparing the generated results with the ground truth, we find that this phenomenon can be attributed to the fact that certain companies within the same industry share identical risks. Consequently, professional analysts include very similar (sometimes identical) risk reminders in their reports. Furthermore, the same company may encounter the same risk in different reports. While the professional opinions may differ in such reports, the risk reminders remain the same. This discrepancy in the risk reminder generation task is the reason why the ROUGE scores and BERT Scores of the experimental models differ significantly from those in the professional opinion generation task.

\subsection{Timing Identification}
In this section, we evaluate the results of the proposed news-triggered opinion expressing timing identification task using macro-averaged F1-score. In addition to using vanilla CPT and BERT as baselines, we employ XL-Sum \cite{hasan-etal-2021-xl} to summarize the given news, and then substitute the generated opinion and risk reminder with the summarization in the proposed approach. Table \ref{tab:Experimental results of timing identification task} illustrates the experimental results. Firstly, comparing the baseline results reveals that CPT performs worse than BERT. Additionally, including the abstract of the news articles leads to a decline in performance. Secondly, we observe that the proposed GADFA demonstrates superior performance in the proposed task regardless of the generator used. Thirdly, the proposed GADFA achieves the best performance when utilizing Mengzi T5 as the generator for generating both professional opinion and risk reminder. Lastly, the results also indicate the importance of timeliness analysis in the task. Incorporating more recent news ($T = 5$) proves to be beneficial in the proposed task.

\begin{table}[t]
  \centering
  \small
    \begin{tabular}{lr}
    \multicolumn{1}{c}{Opinion Generator} & \multicolumn{1}{c}{Macro-F1} \\
    \hline
    mT5  & 67.45\% \\
    Pegasus  & 67.84\% \\
    Mengzi T5  & \textbf{70.58\%} \\
    \end{tabular}%
  \caption{Ablation analysis --- Remove news and risk reminder.}
  \label{tab:blation analysis --- Remove news and risk reminder}%
\end{table}%

\section{Discussion}
\subsection{Ablation Analysis}
In Table~\ref{tab:Experimental results of timing identification task}, we have already demonstrated the effectiveness of adding generated opinion and risk reminder in the proposed task, and we have also shown that models perform better when using $T = 5$.

In this section, we present an ablation analysis of the proposed GADFA by removing the risk reminder from the decision model's input under the $T = 5$ setting. The upper part of Table~\ref{tab:ablation - risk reminder} reveals that the performances deteriorate when using only news and generated opinion. Notably, among these results, we observe that using the professional opinions generated by Pegasus or Mengzi T5 yields better performance compared to using those generated by mT5.

Considering that generators may produce professional opinions from different perspectives, we further explore the scenario of using multiple professional opinions generated by different generators. Specifically, we treat each generator as an independent expert and incorporate the opinions of different experts into the decision-making process. The lower part of Table~\ref{tab:ablation - risk reminder} displays the results of different combinations. We find that utilizing all opinions leads to slightly improved performance. Moreover, the performance achieved by using the generated opinions of Pegasus or Mengzi T5 in conjunction with mT5 surpasses that obtained by using only the generated opinions of mT5.

We further exclude the news from the decision model's input to address the following two research questions: 
(1) Does the decision model still require news for reference despite generating professional opinions based on the given news? and 
(2) Does the performance of the generated results in Table~\ref{tab:Experimental results of generator} reflect the performance in the proposed downstream task? 
Table~\ref{tab:blation analysis --- Remove news and risk reminder} presents the ablation analysis for these two research questions. 
Firstly, we observe that the decision model still relies on news for reference as the performance experiences a significant drop when the news is removed. 
Secondly, we find a positive correlation between the generation performance presented in Table~\ref{tab:Experimental results of generator} and the performance of timing identification. 
This result suggests that there might be a positive correlation between the quality of generation and the performance of the downstream task.

\begin{table}[t]
  \centering
  \small
    \begin{tabular}{l|rr|rr}
          & \multicolumn{2}{c|}{Professional Opinion} & \multicolumn{2}{c}{Risk Reminder} \\
    \cline{2-5}
          & \multicolumn{1}{c}{Sentiment} & \multicolumn{1}{c|}{Entity} & \multicolumn{1}{c}{Sentiment} & \multicolumn{1}{c}{Entity} \\
    \hline
    mT5   & 25.44\% & 0.47\% & 65.68\% & 72.05\% \\
    Pegasus & 48.38\% & 39.96\% & 71.96\% & 72.35\% \\
    Mengzi T5 & 49.38\% & 40.06\% & 76.01\% & 79.50\% \\
    \end{tabular}%
      \caption{Evaluation from sentiment and entity aspects.}
  \label{tab:Evaluation from sentiment and entity aspects}%
\end{table}%

\begin{table}[t]
  \centering
  \resizebox{\columnwidth}{!}{
    \begin{tabular}{l|cc|cc}
          & \multicolumn{2}{c|}{Professional Opinion} & \multicolumn{2}{c}{Risk Reminder} \\
\cline{2-5}          & Sentiment & Entity & Sentiment & Entity \\
    \hline
    mT5 vs. Pegasus & 24.94\% & 25.60\% & 63.47\% & 43.76\% \\
    mT5 vs. Mengzi T5 & 29.68\% & 27.63\% & 64.57\% & 70.69\% \\
    Pegasus vs. Mengzi T5 & 52.37\% & 25.05\% & 59.78\% & 47.77\% \\
    \end{tabular}%
    }
        \caption{Comparison among the generated text of different generators.}
  \label{tab:Comparison among the generated text of different generators}%
\end{table}%

\subsection{Sentiment and Topic}
\label{sec:Sentiment and Topic}
When analyzing opinions, two important aspects need to be considered: sentiment and topic. To evaluate the generated opinions and risk reminders in depth, we employ Stanza~\cite{qi-etal-2020-stanza} and FR2KG~\cite{wang2021data} for sentiment estimation and financial entity extraction, respectively. Stanza is a linguistic analysis toolkit. We obtain the sentiment estimation of both the ground truth and the generated text using Stanza and calculate the ratio of generated text having the same sentiment as the ground truth. FR2KG is a knowledge graph constructed based on Chinese financial reports. We use the entity list in FR2KG to determine the extent to which the generated text contains the same financial entity as the ground truth. If two contents contain the same financial entity, it implies that they are discussing similar topics.

Table~\ref{tab:Evaluation from sentiment and entity aspects} presents the evaluation results. Firstly, we observe that Mengzi T5 performs the best when evaluating both sentiment and entity aspects in the professional opinion generation task. Secondly, in the risk reminder generation task, although Pegasus exhibits the best performance in Table~\ref{tab:Experimental results of generator}, Mengzi T5 outperforms Pegasus in both sentiment and entity aspects. This suggests that one of the possible reasons why the proposed GADFA performs the best is when using Mengzi T5 as the generator for professional opinions and risk reminders. Thirdly, we notice that mT5 performs significantly worse in generating the same entity as the ground truth. Finally, the results align with the findings in Table~\ref{tab:Experimental results of generator}: generating risk reminders appears to be much easier than generating professional opinions.

In addition to comparing the generated text with the ground truth, we provide cross-model comparisons in Table~\ref{tab:Comparison among the generated text of different generators}. Although the generators are trained on the same training set, the sentiment of the generated text varies greatly when given the same news, particularly in the professional opinion generation task. Additionally, Pegasus tends to generate different entities compared to the other two models. This phenomenon is evident in the risk reminder generation task.

\begin{table}[t]
  \centering
  \small
    \begin{tabular}{lr|lr}
    \multicolumn{2}{c|}{Timing - Release} & \multicolumn{2}{c}{View - Change} \\
    \hline
    lift rates  & 1.933 & honeymoon & 2.789 \\
    trade war  & 1.918 & end   & 2.759 \\
    interfere  & 1.892 & slow down  & 2.719 \\
    bulk order  & 1.836 & surprise  & 2.567 \\
    exchange rate  & 1.827 & gap   & 2.496 \\
    \end{tabular}%
        \caption{Keywords selected based on PMI score.}
  \label{tab:Keywords selected based on PMI. (Release Report and Change View)}%
\end{table}%

\begin{table*}[t]
  \centering
   \resizebox{\textwidth}{!}{
    \begin{tabular}{l|p{53em}}
    \multirow{2}[0]{*}{Given News}  & Uni-President announced its Q3 financial report last night. The net profit after tax in the third quarter was \textbf{5.032 billion yuan}, down 8.23\% quarter-on-quarter and 11.92\% year-on-year. The single-quarter EPS was 0.89 yuan, a relatively large decline. \\
    \hline
    \multirow{5}[0]{*}{Analyst} & Third-quarter earnings fell short of expectations, mainly due to slower earnings growth at several key subsidiaries. We continue to view UPC's high-end product strategy positively, coupled with a 3\% cash yield, which is usually a safe haven during market volatility. However, it is believed that the current stock price upside is limited, mainly due to the slow growth of the Chinese business entering the off-season and the slow growth of convenience stores due to e-commerce fees. It is recommended to be neutral, and the EPS in \textbf{2018/19} is estimated to be \textbf{3.08/3.17 yuan}. \\
    \hline
    \multirow{3}[0]{*}{mT5}   & The Center believes that the decline in UPC's net profit after tax in the fourth quarter is expected to increase quarter by quarter, because 1) the Chinese government intends to reduce the pressure through early repayment and lower interest rates. Stress is lessened, so we remain neutral on Uni-President Enterprises Corporation. \\
    \hline
    \multirow{5}[0]{*}{Pegasus} & Uni-President's 3Q19 net profit after tax was \textbf{99 billion yuan}, lower than expected, mainly due to the group's deferred capital expenditures and shipments. Looking forward to the market outlook, we believe that Uni-President will continue to benefit from its high-value strategy, coupled with the strengthening of Uni-President's profit rate outlook, and Uni-President's same-store sales growth in the second half of 2019 is expected to improve, and we recommend paying close attention. Estimated EPS for \textbf{2019/20} are \textbf{3.44/3.68 yuan}. \\
    \hline
    \multirow{5}[0]{*}{Mengzi T5} & Although Uni-President's revenue showed a quarterly decline, Uni-President's revenue showed a quarter-to-quarter decline that was slightly lower than expected. We continue to take a positive view on UPC's high-end product strategy and the recent purchases on revenue and profit. However, we suggest waiting for a better time, because: 1) China's business is facing slower consumption upgrades and intensified competition; 2) UPC has surpassed this year's Earnings outlook is flat, as e-commerce fee revenue growth momentum slows and reinvestment gains are modest. \\
    \end{tabular}%
    }
      \caption{Examples of generated opinions.}
  \label{tab:Examples of generated opinions}%
\end{table*}%

\subsection{Analysis of Professionals' and Generated Opinions}
To provide a more in-depth analysis of the events that trigger professionals' behaviors, we employ pointwise mutual information (PMI) to calculate word-level scores\footnote{We will release the wordlist and the PMI scores.}. PMI is a widely used method for constructing sentiment dictionaries~\cite{khan2016sentimi}, and we believe it can also provide valuable insights for mining professionals' behaviors. 
Apart from considering the timing of report releases, we conduct additional analyses to comprehend the factors that prompt professional analysts to alter their perspective on a given stock. 
The PMI score of a word is computed as:
\begin{equation}
    \log_2 \frac{p(w,\mathit{condition})}{p(w)p(\mathit{condition})} - \log_2 \frac{p(w,\neg \mathit{condition})}{p(w)p(\neg\mathit{condition})},
\end{equation}
Here, $w$ represents the target word in the provided news, while $\mathit{condition}$ denotes analysts' behaviors, such as releasing reports or changing their view.

Table~\ref{tab:Keywords selected based on PMI. (Release Report and Change View)} presents the statistical results. Firstly, we observe a strong correlation between the timing of report releases and macroeconomic events such as lift rates, trade wars, exchange rates, as well as significant company news like bulk orders. Secondly, we find that changes in views are primarily influenced by changes in status (ending or slowing down) and unexpected events (surprises or gaps). It is important to note that "gap" refers to a substantial rise or fall in stock prices or earnings.

In Table~\ref{tab:Examples of generated opinions}, we provide an illustrative example demonstrating the models' proficiency in generating coherent and plausible investment opinions. Furthermore, both the generation models and professional analysts express neutral sentiments towards the given company. However, there exist some nuanced inaccuracies, which in turn raise additional research questions for future investigations. Firstly, the models may generate false numerical information, as evidenced by the provided news and Pegasus's result (5.032 billion vs. 99 billion). Secondly, the temporal information in the generated opinions may differ from that expressed by analysts, as highlighted by the analyst's opinion and Pegasus's result (2018/19 vs. 2019/20). Thirdly, estimations differ between analysts' opinions and generated opinions, as exemplified by the analyst's opinion and Pegasus's result (3.08/3.17 yuan vs. 3.44/3.68 yuan). These findings prompt the following research questions for future investigations:

\begin{itemize}
    \item How can we generate accurate numerical information?
    
    \item Should the generator be considered faulty when the generated estimation deviates from that of professional analysts?
    
    \item Does numerical information influence the performance of the decision model?
\end{itemize}

To further analyze the model properties, we present the statistics of the generated results in Table~\ref{tab:Statistics of generated results}. Firstly, the models generate shorter texts than the ground truth for both investment generation and risk reminder generation tasks. Secondly, the models utilize fewer tokens compared to the ground truth, with mT5 employing the least tokens among the three models. Thirdly, approximately 50\% of words in risk reminders can be directly copied from the input, which is a professional opinion. This is reasonable since the risk of the professional opinion, which is based on the expectation of sales growth, would be an event that slows down the sales growth. For instance, the risk reminder for the professional opinion "Multiple high-end new \textit{drivers} can grow significantly on Silergy’s improving product offerings and project wins" is "Slower progress in any of the key growth \textit{drivers}".

Finally, we analyze the distributions of the sentiment using the same tool, Stanza, in Section~\ref{sec:Sentiment and Topic}. We observe that Pegasus's sentiment distribution in the professional opinion generation task is much closer to the ground truth's distribution, while Mengzi T5's sentiment distribution in the risk reminder generation task is also much closer to the ground truth's distribution. Unlike the other two models, mT5 generates a higher proportion of neutral professional opinions. By comparing the statistics of the ground truth in both tasks, we discover that analysts tend to write more positive reports than negative reports. Moreover, risk reminders are predominantly written in a negative tone rather than a positive tone.

\begin{table}[t]
  \centering
  \resizebox{\columnwidth}{!}{
    \begin{tabular}{ll|r|r}
          &       & \multicolumn{1}{c|}{Professional Opinion} & \multicolumn{1}{c}{Risk Reminder} \\
    \hline
    \multirow{6}[2]{*}{Analyst} & Avg. \# Words & 93.29 & 44.56 \\
          & Unique Token &                            15,217  &                 2,113  \\
          & Words from Input & 15.33\% & 49.60\% \\
          & Positive  & 46.88\% & 20.66\% \\
          & Negative & 33.17\% & 49.08\% \\
          & Neutral & 19.95\% & 30.26\% \\
    \hline
    \multirow{6}[2]{*}{mT5} & Avg. \# Words & 66.23 & 37.77 \\
          & Unique Token &                              1,873  & 915 \\
          & Words from Input & 33.68\% & 72.54\% \\
          & Positive  & 12.72\% & 21.40\% \\
          & Negative & 3.24\% & 49.08\% \\
          & Neutral & 84.04\% & 29.52\% \\
    \hline
    \multirow{6}[2]{*}{Pegasus} & Avg. \# Words & 63.53 & 40.20 \\
          & Unique Token &                              3,865  &                 1,110  \\
          & Words from Input & 32.91\% & 74.88\% \\
          & Positive  & 46.88\% & 17.71\% \\
          & Negative & 30.92\% & 35.43\% \\
          & Neutral & 22.19\% & 46.86\% \\
    \hline
    \multirow{6}[1]{*}{Mengzi T5} & Avg. \# Words & 68.08 & 38.82 \\
          & Unique Token &                              3,913  &                 1,054  \\
          & Words from Input & 31.76\% & 75.41\% \\
          & Positive  & 42.14\% & 19.56\% \\
          & Negative & 30.42\% & 50.55\% \\
          & Neutral & 27.43\% & 29.89\% \\
    \end{tabular}%
    }
        \caption{Statistics of generated results. The input of professional opinions is news, and the input of risk reminders is professional opinions.}
  \label{tab:Statistics of generated results}%
\end{table}%

\subsection{Distribution of Report Release Date}
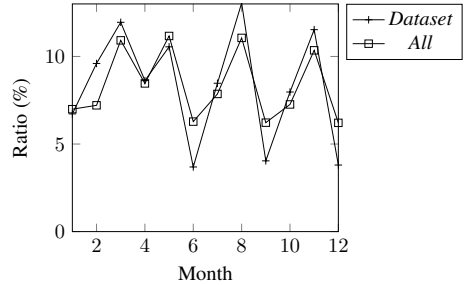
\begin{figure}[t]
\centering
\pgfplotsset{width=5cm,compat=1.14}
\begin{tikzpicture}[scale=0.7]
\begin{axis}[
  scale only axis,
  legend pos=outer north east,
  ymin=0,
  ymax=13,
  xmin=1,
  xmax=12,
  xlabel=Month,
  ylabel=Ratio (\%),
]
  \addplot[mark=+]
    coordinates{
    (1,6.72)
    (2,9.60)
    (3,11.95)
    (4,8.62)
    (5,10.55)
    (6,3.69)
    (7,8.47)
    (8,13.05)
    (9,4.04)
    (10,7.97)
    (11,11.53)
    (12,3.8)
    }; \label{plot_1_y1}
  \addlegendentry{\textit{Dataset}}

  \addplot[mark=square]
    coordinates{
    (1,6.99)
    (2,7.21)
    (3,10.92)
    (4,8.46)
    (5,11.17)
    (6,6.28)
    (7,7.86)
    (8,11.06)
    (9,6.22)
    (10,7.26)
    (11,10.35)
    (12,6.21)
    }; \label{plot_1_y2}
    \addlegendentry{\textit{All}}
 
\addlegendimage{/pgfplots/refstyle=plot_1_y1}
\addlegendimage{/pgfplots/refstyle=plot_1_y2}
 
\end{axis}
\end{tikzpicture}

\caption{Distribution of report release date within year. Dataset and All denote the statistics of the experimental dataset and all reports from 2014 to 2020, respectively.}
\label{fig:Distribution of report release date - year}
\end{figure}

This section presents statistics on the distribution of report release dates. Figure~\ref{fig:Distribution of report release date - year} illustrates the statistics within the year. In addition to the statistics based on the experimental dataset, we include the statistics based on all reports from 2014 to 2020 for comparison. Firstly, the distributions based on the experimental dataset and all reports are similar. Secondly, we observe that analysts tend to release more reports in March, August, and November, while fewer reports are released in June, September, and December. One reason for the increased number of reports in March, August, and November is that companies are required to release their yearly, second-quarter, and third-quarter financial statements before the end of these months. The first-quarter financial statement is expected to be released before the end of May, resulting in the fourth-highest number of reports in May. Since professional analysts do not frequently change their views, the months following the financial statement release dates, i.e., June, September, and December, have fewer reports.

Table~\ref{tab:Distribution of report release date - month} presents the distribution of report release dates within a month. It is observed that analysts release fewer reports at the beginning of the month, while the ratios at the middle and end of the month are similar. This trend can be attributed to the requirement for companies to disclose the previous month's revenue before the 10th of each month, with most companies releasing this information around that time. Consequently, professionals tend to release reports during the middle of the month to incorporate the latest information into their opinions. Nevertheless, there are still instances of reports being released at other times. The proposed task and dataset serve as a testbed for identifying the optimal timing of expressing an opinion based on news information. The provided approach and dataset offer avenues for future research, such as enhancing the performance of the proposed GADFA or developing a new end-to-end model to address the proposed task.

\begin{table}[t]
  \centering
  \small
    \begin{tabular}{lrr}
    \multicolumn{1}{c}{Period} & \multicolumn{1}{c}{Dataset} & \multicolumn{1}{c}{All} \\
    \hline
    Beginning of the month & 29.04\% & 29.73\% \\
    Middle of the month   & 34.48\% & 35.41\% \\
    End of the month  & 36.47\% & 34.85\% \\
    \end{tabular}%
        \caption{Distribution of release date within month.}
  \label{tab:Distribution of report release date - month}%
\end{table}%

\section{Conclusion}
We introduced the task of identifying the timing of expressing opinions triggered by news and proposed GADFA to address this task. We demonstrated that augmenting the decision model with risk reminders improves performance. Various discussions on model comparison, linguistic features, and analyst behavior based on temporal conditions provided a comprehensive understanding of GADFA and the task.
Our approach is extendable to other professional behavior forecasting tasks, aligning with the steps involved in the decision-making process. In the future, we plan to explore GADFA's application in domains like clinical decision-making and forecasting trading behavior.

\section*{Limitations}
The limitations of this paper are four-fold as follows:
\begin{enumerate}
    \item Due to the data availability, this paper only explores the proposed task with Chinese data, and thus cannot claim that the proposed GADFA is a general approach to any languages. 
    Future work can use the proposed approach in other languages' application scenarios. 

    \item As we mentioned, experts in other fields, such as CDC's experts and politicians, also need to identify the timing of expressing opinions. 
    However, this paper only explores one application scenario, and cannot claim that the proposed approach is also useful in other scenarios. 
    We believe that this paper provides a comprehensive analysis of analysts' opinion expression timing. 
    Future works can follow our line of thought to explore the proposed task in other domains. 
    
    \item Since this paper focuses on the timing identification task, we mainly pay attention to the performance of the proposed task. 
    Although we analyzed the generation results with ROUGE scores, BERT Scores, sentiment, and entity aspects, we did not perform the human evaluation on the generated results. 
    Because the performances of the generator used in the experiments were already verified with human evaluation when they were proposed, we did not do human evaluation again. 
    We focused on how these models influence the performance of the proposed task.
    If future work attempt to improve the performance by proposing a new generator, we think that it would be great to compare generators with human evaluation.
    
    \item The released dataset can only be used in academic usage. 
    In the proposed dataset, we share the news and corresponding labels on the timing identification task. 
    Regarding news-opinion and risk-reminder pairs, future works can get them by applying them to the authors. 
    The applicant must sign the agreement of the exclusive usage and agree to follow the CC BY-NC-SA 4.0 license. 
    For industrial usage, we can guide the researcher to rebuild our dataset from different information vendors.

\end{enumerate}

\section*{Impact Statement}
Although this paper can help go one more step further in opinion mining and automatic stock analysis assistance, some potential negative outcomes exist if developers use our method with bad intent~\cite{solaiman2019release}. 
For example, the NLP-based trading algorithm would easily be influenced by changing one word~\cite{xie2022advtweet}. 
We think that the proposed timing identification task can also be used to influence the financial market.
For example, the developer can select some timing to release generated opinions with false information, and it may lead to large market volatility. 
However, we believe that understanding the task and models' properties in-depth can also help us avoid this potential risk.

\bibliography{custom}

\begin{thebibliography}{44}
\expandafter\ifx\csname natexlab\endcsname\relax\def\natexlab#1{#1}\fi

\bibitem[{Black and Scholes(1973)}]{black1973pricing}
Fischer Black and Myron Scholes. 1973.
\newblock The pricing of options and corporate liabilities.
\newblock \emph{The Journal of Political Economy}, 81(3):637--654.

\bibitem[{Chen et~al.(2022)Chen, Teng, Wang, and Zhang}]{chen-etal-2022-discrete}
Chenhua Chen, Zhiyang Teng, Zhongqing Wang, and Yue Zhang. 2022.
\newblock \href {https://doi.org/10.18653/v1/2022.acl-long.145} {Discrete opinion tree induction for aspect-based sentiment analysis}.
\newblock In \emph{Proceedings of the 60th Annual Meeting of the Association for Computational Linguistics}, pages 2051--2064, Dublin, Ireland. Association for Computational Linguistics.

\bibitem[{Chuang and Yang(2022)}]{chuang-yang-2022-buy}
Chengyu Chuang and Yi~Yang. 2022.
\newblock \href {https://doi.org/10.18653/v1/2022.acl-short.12} {Buy tesla, sell ford: Assessing implicit stock market preference in pre-trained language models}.
\newblock In \emph{Proceedings of the 60th Annual Meeting of the Association for Computational Linguistics (Volume 2: Short Papers)}, pages 100--105, Dublin, Ireland. Association for Computational Linguistics.

\bibitem[{Conrad et~al.(2006)Conrad, Cornell, Landsman, and Rountree}]{conrad2006analyst}
Jennifer Conrad, Bradford Cornell, Wayne~R Landsman, and Brian~R Rountree. 2006.
\newblock How do analyst recommendations respond to major news?
\newblock \emph{Journal of Financial and Quantitative Analysis}, 41(1):25--49.

\bibitem[{Devlin et~al.(2019)Devlin, Chang, Lee, and Toutanova}]{devlin-etal-2019-bert}
Jacob Devlin, Ming-Wei Chang, Kenton Lee, and Kristina Toutanova. 2019.
\newblock \href {https://doi.org/10.18653/v1/N19-1423} {{BERT}: Pre-training of deep bidirectional transformers for language understanding}.
\newblock In \emph{Proceedings of the 2019 Conference of the North {A}merican Chapter of the Association for Computational Linguistics: Human Language Technologies, Volume 1 (Long and Short Papers)}, pages 4171--4186, Minneapolis, Minnesota. Association for Computational Linguistics.

\bibitem[{Devos et~al.(2015)Devos, Hao, Prevost, and Wongchoti}]{devos2015stock}
Erik Devos, Wei Hao, Andrew~K Prevost, and Udomsak Wongchoti. 2015.
\newblock Stock return synchronicity and the market response to analyst recommendation revisions.
\newblock \emph{Journal of Banking \& Finance}, 58:376--389.

\bibitem[{Diaz and Ng(2018)}]{diaz2018modeling}
Gerardo~Ocampo Diaz and Vincent Ng. 2018.
\newblock Modeling and prediction of online product review helpfulness: a survey.
\newblock In \emph{Proceedings of the 56th Annual Meeting of the Association for Computational Linguistics}, pages 698--708.

\bibitem[{Fama(1995)}]{fama1995random}
Eugene~F Fama. 1995.
\newblock Random walks in stock market prices.
\newblock \emph{Financial analysts journal}, 51(1):75--80.

\bibitem[{Hasan et~al.(2021)Hasan, Bhattacharjee, Islam, Mubasshir, Li, Kang, Rahman, and Shahriyar}]{hasan-etal-2021-xl}
Tahmid Hasan, Abhik Bhattacharjee, Md.~Saiful Islam, Kazi Mubasshir, Yuan-Fang Li, Yong-Bin Kang, M.~Sohel Rahman, and Rifat Shahriyar. 2021.
\newblock {XL}-sum: Large-scale multilingual abstractive summarization for 44 languages.
\newblock In \emph{ACL-IJCNLP}.

\bibitem[{Hirst et~al.(1995)Hirst, Koonce, and Simko}]{hirst1995investor}
D~Eric Hirst, Lisa Koonce, and Paul~J Simko. 1995.
\newblock Investor reactions to financial analysts' research reports.
\newblock \emph{Journal of Accounting Research}, 33(2):335--351.

\bibitem[{Hsieh et~al.(2016)Hsieh, Hui, and Zhang}]{hsieh2016analyst}
Chia-Chun Hsieh, Kai~Wai Hui, and Yao Zhang. 2016.
\newblock Analyst report readability and stock returns.
\newblock \emph{Journal of Business Finance \& Accounting}, 43(1-2):98--130.

\bibitem[{Hsu and Tan(2021)}]{hsu-tan-2021-decision}
Chao-Chun Hsu and Chenhao Tan. 2021.
\newblock \href {https://doi.org/10.18653/v1/2021.emnlp-main.10} {Decision-focused summarization}.
\newblock In \emph{Proceedings of the 2021 Conference on Empirical Methods in Natural Language Processing}, pages 117--132, Online and Punta Cana, Dominican Republic. Association for Computational Linguistics.

\bibitem[{Hua et~al.(2019)Hua, Hu, and Wang}]{hua-etal-2019-argument-generation}
Xinyu Hua, Zhe Hu, and Lu~Wang. 2019.
\newblock \href {https://doi.org/10.18653/v1/P19-1255} {Argument generation with retrieval, planning, and realization}.
\newblock In \emph{Proceedings of the 57th Annual Meeting of the Association for Computational Linguistics}, pages 2661--2672, Florence, Italy. Association for Computational Linguistics.

\bibitem[{Keith and Stent(2019)}]{keith-stent-2019-modeling}
Katherine Keith and Amanda Stent. 2019.
\newblock \href {https://doi.org/10.18653/v1/P19-1047} {Modeling financial analysts{'} decision making via the pragmatics and semantics of earnings calls}.
\newblock In \emph{Proceedings of the 57th Annual Meeting of the Association for Computational Linguistics}, pages 493--503, Florence, Italy. Association for Computational Linguistics.

\bibitem[{Khan et~al.(2016)Khan, Qamar, and Bashir}]{khan2016sentimi}
Farhan~Hassan Khan, Usman Qamar, and Saba Bashir. 2016.
\newblock Sentimi: Introducing point-wise mutual information with sentiwordnet to improve sentiment polarity detection.
\newblock \emph{Applied Soft Computing}, 39:140--153.

\bibitem[{Kim and Ryu(2022)}]{kim2022firm}
Yongsik Kim and Doojin Ryu. 2022.
\newblock Firm-specific or market-wide information: How does analyst coverage influence stock price synchronicity?
\newblock \emph{Available at SSRN 4010314}.

\bibitem[{Kothari et~al.(2016)Kothari, So, and Verdi}]{kothari2016analysts}
Sagar~P Kothari, Eric So, and Rodrigo Verdi. 2016.
\newblock Analysts’ forecasts and asset pricing.
\newblock \emph{Annual Review of Financial Economics}, 8:197--219.

\bibitem[{Li et~al.(2020{\natexlab{a}})Li, Yang, Smyth, and Dong}]{li2020maec}
Jiazheng Li, Linyi Yang, Barry Smyth, and Ruihai Dong. 2020{\natexlab{a}}.
\newblock Maec: A multimodal aligned earnings conference call dataset for financial risk prediction.
\newblock In \emph{Proceedings of the 29th ACM International Conference on Information \& Knowledge Management}, pages 3063--3070.

\bibitem[{Li et~al.(2020{\natexlab{b}})Li, Bao, Harimoto, Chen, Xu, and Su}]{li2020modeling}
Wei Li, Ruihan Bao, Keiko Harimoto, Deli Chen, Jingjing Xu, and Qi~Su. 2020{\natexlab{b}}.
\newblock Modeling the stock relation with graph network for overnight stock movement prediction.
\newblock In \emph{IJCAI}, volume~20, pages 4541--4547.

\bibitem[{Lin(2004)}]{lin-2004-rouge}
Chin-Yew Lin. 2004.
\newblock \href {https://aclanthology.org/W04-1013} {{ROUGE}: A package for automatic evaluation of summaries}.
\newblock In \emph{Text Summarization Branches Out}, pages 74--81, Barcelona, Spain. Association for Computational Linguistics.

\bibitem[{Ling et~al.(2022)Ling, Yu, and Xia}]{ling-etal-2022-vision}
Yan Ling, Jianfei Yu, and Rui Xia. 2022.
\newblock \href {https://doi.org/10.18653/v1/2022.acl-long.152} {Vision-language pre-training for multimodal aspect-based sentiment analysis}.
\newblock In \emph{Proceedings of the 60th Annual Meeting of the Association for Computational Linguistics}, pages 2149--2159, Dublin, Ireland. Association for Computational Linguistics.

\bibitem[{Liu(2012)}]{liu2012sentiment}
Bing Liu. 2012.
\newblock Sentiment analysis and opinion mining.
\newblock \emph{Synthesis lectures on human language technologies}, 5(1):1--167.

\bibitem[{Mandi et~al.(2022)Mandi, Bucarey, Tchomba, and Guns}]{mandi2022decision}
Jayanta Mandi, Victor Bucarey, Maxime Mulamba~Ke Tchomba, and Tias Guns. 2022.
\newblock Decision-focused learning: Through the lens of learning to rank.
\newblock In \emph{International Conference on Machine Learning}, pages 14935--14947. PMLR.

\bibitem[{Niehaus and Zhang(2010)}]{niehaus2010impact}
Greg Niehaus and Donghang Zhang. 2010.
\newblock The impact of sell-side analyst research coverage on an affiliated broker’s market share of trading volume.
\newblock \emph{Journal of Banking \& Finance}, 34(4):776--787.

\bibitem[{Qi et~al.(2020)Qi, Zhang, Zhang, Bolton, and Manning}]{qi-etal-2020-stanza}
Peng Qi, Yuhao Zhang, Yuhui Zhang, Jason Bolton, and Christopher~D. Manning. 2020.
\newblock \href {https://doi.org/10.18653/v1/2020.acl-demos.14} {{S}tanza: A python natural language processing toolkit for many human languages}.
\newblock In \emph{Proceedings of the 58th Annual Meeting of the Association for Computational Linguistics: System Demonstrations}, pages 101--108, Online. Association for Computational Linguistics.

\bibitem[{Qin and Yang(2019)}]{qin-yang-2019-say}
Yu~Qin and Yi~Yang. 2019.
\newblock \href {https://doi.org/10.18653/v1/P19-1038} {What you say and how you say it matters: Predicting stock volatility using verbal and vocal cues}.
\newblock In \emph{Proceedings of the 57th Annual Meeting of the Association for Computational Linguistics}, pages 390--401, Florence, Italy. Association for Computational Linguistics.

\bibitem[{Raffel et~al.(2020)Raffel, Shazeer, Roberts, Lee, Narang, Matena, Zhou, Li, Liu et~al.}]{raffel2020exploring}
Colin Raffel, Noam Shazeer, Adam Roberts, Katherine Lee, Sharan Narang, Michael Matena, Yanqi Zhou, Wei Li, Peter~J Liu, et~al. 2020.
\newblock Exploring the limits of transfer learning with a unified text-to-text transformer.
\newblock \emph{J. Mach. Learn. Res.}, 21(140):1--67.

\bibitem[{Sawhney et~al.(2021)Sawhney, Aggarwal, and Shah}]{sawhney-etal-2021-empirical}
Ramit Sawhney, Arshiya Aggarwal, and Rajiv~Ratn Shah. 2021.
\newblock \href {https://doi.org/10.18653/v1/2021.naacl-main.294} {An empirical investigation of bias in the multimodal analysis of financial earnings calls}.
\newblock In \emph{Proceedings of the 2021 Conference of the North American Chapter of the Association for Computational Linguistics: Human Language Technologies}, pages 3751--3757, Online. Association for Computational Linguistics.

\bibitem[{Schiller et~al.(2021)Schiller, Daxenberger, and Gurevych}]{schiller-etal-2021-aspect}
Benjamin Schiller, Johannes Daxenberger, and Iryna Gurevych. 2021.
\newblock \href {https://doi.org/10.18653/v1/2021.naacl-main.34} {Aspect-controlled neural argument generation}.
\newblock In \emph{Proceedings of the 2021 Conference of the North American Chapter of the Association for Computational Linguistics: Human Language Technologies}, pages 380--396, Online. Association for Computational Linguistics.

\bibitem[{Shao et~al.(2021)Shao, Geng, Liu, Dai, Yang, Zhe, Bao, and Qiu}]{shao2021cpt}
Yunfan Shao, Zhichao Geng, Yitao Liu, Junqi Dai, Fei Yang, Li~Zhe, Hujun Bao, and Xipeng Qiu. 2021.
\newblock {CPT}: A pre-trained unbalanced transformer for both chinese language understanding and generation.
\newblock \emph{arXiv preprint arXiv:2109.05729}.

\bibitem[{Shi et~al.(2022)Shi, Li, Li, Fei, and Ji}]{shi-etal-2022-effective}
Wenxuan Shi, Fei Li, Jingye Li, Hao Fei, and Donghong Ji. 2022.
\newblock \href {https://doi.org/10.18653/v1/2022.acl-long.291} {Effective token graph modeling using a novel labeling strategy for structured sentiment analysis}.
\newblock In \emph{Proceedings of the 60th Annual Meeting of the Association for Computational Linguistics}, pages 4232--4241, Dublin, Ireland. Association for Computational Linguistics.

\bibitem[{Solaiman et~al.(2019)Solaiman, Brundage, Clark, Askell, Herbert-Voss, Wu, Radford, Krueger, Kim, Kreps et~al.}]{solaiman2019release}
Irene Solaiman, Miles Brundage, Jack Clark, Amanda Askell, Ariel Herbert-Voss, Jeff Wu, Alec Radford, Gretchen Krueger, Jong~Wook Kim, Sarah Kreps, et~al. 2019.
\newblock Release strategies and the social impacts of language models.
\newblock \emph{arXiv preprint arXiv:1908.09203}.

\bibitem[{Wachsmuth et~al.(2018)Wachsmuth, Stede, El~Baff, Al-Khatib, Skeppstedt, and Stein}]{wachsmuth-etal-2018-argumentation}
Henning Wachsmuth, Manfred Stede, Roxanne El~Baff, Khalid Al-Khatib, Maria Skeppstedt, and Benno Stein. 2018.
\newblock \href {https://aclanthology.org/C18-1318} {Argumentation synthesis following rhetorical strategies}.
\newblock In \emph{Proceedings of the 27th International Conference on Computational Linguistics}, pages 3753--3765, Santa Fe, New Mexico, USA. Association for Computational Linguistics.

\bibitem[{Wang et~al.(2021)Wang, Xu, Du, Chen, Wang, and Wen}]{wang2021data}
Wenguang Wang, Yonglin Xu, Chunhui Du, Yunwen Chen, Yijie Wang, and Hui Wen. 2021.
\newblock Data set and evaluation of automated construction of financial knowledge graph.
\newblock \emph{Data Intelligence}, 3(3):418--443.

\bibitem[{Wilder et~al.(2019)Wilder, Dilkina, and Tambe}]{wilder2019melding}
Bryan Wilder, Bistra Dilkina, and Milind Tambe. 2019.
\newblock Melding the data-decisions pipeline: Decision-focused learning for combinatorial optimization.
\newblock In \emph{Proceedings of the AAAI Conference on Artificial Intelligence}, volume~33, pages 1658--1665.

\bibitem[{Wolf et~al.(2019)Wolf, Debut, Sanh, Chaumond, Delangue, Moi, Cistac, Rault, Louf, Funtowicz et~al.}]{wolf2019huggingface}
Thomas Wolf, Lysandre Debut, Victor Sanh, Julien Chaumond, Clement Delangue, Anthony Moi, Pierric Cistac, Tim Rault, R{\'e}mi Louf, Morgan Funtowicz, et~al. 2019.
\newblock Huggingface's transformers: State-of-the-art natural language processing.
\newblock \emph{arXiv preprint arXiv:1910.03771}.

\bibitem[{Wu and Shi(2022)}]{wu-shi-2022-adversarial}
Hui Wu and Xiaodong Shi. 2022.
\newblock \href {https://doi.org/10.18653/v1/2022.acl-long.174} {Adversarial soft prompt tuning for cross-domain sentiment analysis}.
\newblock In \emph{Proceedings of the 60th Annual Meeting of the Association for Computational Linguistics}, pages 2438--2447, Dublin, Ireland. Association for Computational Linguistics.

\bibitem[{Xie et~al.(2022)Xie, Wang, Chen, Jinjun, Liu, and Koyejo}]{xie2022advtweet}
Yong Xie, Dakuo Wang, Pin-Yu Chen, Xiong Jinjun, Sijia Liu, and Oluwasanmi Koyejo. 2022.
\newblock A word is worth a thousand dollars: Adversarial attack on tweets fools stock prediction.
\newblock \emph{Proceedings of the 2022 Conference of the North American Chapter of the Association for Computational Linguistics: Human Language Technologies}.

\bibitem[{Xu and Cohen(2018)}]{xu-cohen-2018-stock}
Yumo Xu and Shay~B. Cohen. 2018.
\newblock \href {https://doi.org/10.18653/v1/P18-1183} {Stock movement prediction from tweets and historical prices}.
\newblock In \emph{Proceedings of the 56th Annual Meeting of the Association for Computational Linguistics}, pages 1970--1979, Melbourne, Australia. Association for Computational Linguistics.

\bibitem[{Xue et~al.(2021)Xue, Constant, Roberts, Kale, Al-Rfou, Siddhant, Barua, and Raffel}]{xue-etal-2021-mt5}
Linting Xue, Noah Constant, Adam Roberts, Mihir Kale, Rami Al-Rfou, Aditya Siddhant, Aditya Barua, and Colin Raffel. 2021.
\newblock \href {https://doi.org/10.18653/v1/2021.naacl-main.41} {m{T}5: A massively multilingual pre-trained text-to-text transformer}.
\newblock In \emph{Proceedings of the 2021 Conference of the North American Chapter of the Association for Computational Linguistics: Human Language Technologies}, pages 483--498, Online. Association for Computational Linguistics.

\bibitem[{Zhang et~al.(2020)Zhang, Zhao, Saleh, and Liu}]{zhang2020pegasus}
Jingqing Zhang, Yao Zhao, Mohammad Saleh, and Peter Liu. 2020.
\newblock Pegasus: Pre-training with extracted gap-sentences for abstractive summarization.
\newblock In \emph{International Conference on Machine Learning}, pages 11328--11339. PMLR.

\bibitem[{Zhang et~al.(2019)Zhang, Kishore, Wu, Weinberger, and Artzi}]{zhang2019bertscore}
Tianyi Zhang, Varsha Kishore, Felix Wu, Kilian~Q Weinberger, and Yoav Artzi. 2019.
\newblock Bertscore: Evaluating text generation with bert.
\newblock In \emph{International Conference on Learning Representations}.

\bibitem[{Zhang et~al.(2021)Zhang, Zhang, Chen, Guo, Hua, Wang, and Zhou}]{zhang2021mengzi}
Zhuosheng Zhang, Hanqing Zhang, Keming Chen, Yuhang Guo, Jingyun Hua, Yulong Wang, and Ming Zhou. 2021.
\newblock \href {http://arxiv.org/abs/2110.06696} {Mengzi: Towards lightweight yet ingenious pre-trained models for chinese}.

\bibitem[{Zhao et~al.(2019)Zhao, Chen, Zhang, Zhao, Liu, Lu, Chen, Deng, Ju, and Du}]{zhao2019uer}
Zhe Zhao, Hui Chen, Jinbin Zhang, Xin Zhao, Tao Liu, Wei Lu, Xi~Chen, Haotang Deng, Qi~Ju, and Xiaoyong Du. 2019.
\newblock Uer: An open-source toolkit for pre-training models.
\newblock \emph{EMNLP-IJCNLP 2019}, page 241.

\end{thebibliography}

\appendix

\begin{table*}[t]
  \centering
  \resizebox{\textwidth}{!}{
    \begin{tabular}{ll}
    \hline
          & \multicolumn{1}{c}{URL} \\
    \hline
    BERT~\cite{devlin-etal-2019-bert} & \url{https://huggingface.co/bert-base-chinese} 
    \\
    CPT~\cite{shao2021cpt} & \url{https://huggingface.co/fnlp/cpt-base}
    \\
    XL-Sum~\cite{hasan-etal-2021-xl} & \url{https://huggingface.co/spaces/krrishD/csebuetnlp_mT5_multilingual_XLSum}
    \\
    mT5~\cite{raffel2020exploring} & \url{https://huggingface.co/google/mt5-base} \\
    Pegasus~\cite{zhang2020pegasus} & \url{https://huggingface.co/uer/pegasus-base-chinese-cluecorpussmall} 
    \\
    Mengzi T5~\cite{zhang2021mengzi} & \url{https://huggingface.co/Langboat/mengzi-t5-base}
    \\
    \hline
    \end{tabular}%
    }
  \caption{Reference for the models in our experiments.}
  \label{tab:Reference for the models in our experiments}%
\end{table*}%

\section{Implement Detail}
\label{sec:Implement Detail}
In our experiment, we use the Hugging Face transformers package~\cite{wolf2019huggingface}. \footnote{\url{https://huggingface.co/docs/transformers/index}}
Intel Xeon Gold CPU and Nvidia Tesla V100 w/32GB are the CPU and GPU we used. 
Table~\ref{tab:Reference for the models in our experiments} provides the links to the models we used in the experiments.

We will release these datasets for academic use under the CC BY-NC-SA 4.0 license.

\end{document}